\documentclass[11pt,letterpaper]{article}
\usepackage{hyperref}
\usepackage{ijcnlp2017}
\usepackage{times}
\usepackage{latexsym}
\usepackage{mathtools}
\usepackage{graphics}
\usepackage{amsmath}
\usepackage{multirow}
\usepackage{url}
\usepackage[utf8]{vietnam}
\usepackage[english]{babel}
\usepackage{listings}

\ijcnlpfinalcopy

\def\ijcnlppaperid{27}

\title{NNVLP: A Neural Network-Based \\Vietnamese Language Processing Toolkit}

\author{Thai-Hoang Pham \\ Alt Inc \\ Hanoi, Vietnam \\ \url{phamthaihoang.hn@gmail.com}
        \And
        Xuan-Khoai Pham \\ FPT University \\ Hanoi, Vietnam \\ \url{khoaipxmse0060@fpt.edu.vn}
        \AND
        Tuan-Anh Nguyen \\ Alt Inc \\ Hanoi, Vietnam \\ \url{ntanh.hus@gmail.com}
        \And
        Phuong Le-Hong \\ Vietnam National University \\ Hanoi, Vietnam \\ \url{phuonglh@vnu.edu.vn}       
        }

\date{}

\begin{document}

\maketitle

\begin{abstract}
This paper demonstrates neural network-based toolkit namely NNVLP for essential
  Vietnamese language processing tasks including part-of-speech (POS) tagging, chunking, named entity recognition (NER). Our toolkit is a
  combination of bidirectional Long Short-Term Memory (Bi-LSTM),
  Convolutional Neural Network (CNN), Conditional Random Field (CRF),
  using pre-trained word embeddings as input, which achieves state-of-the-art results on these three tasks.  We provide both API and web demo\footnote{\url{nnvlp.org}} for this toolkit.
\end{abstract}

\section{Introduction}
\label{sec:introduction}
\paragraph{}
Vietnamese belongs to the top 20 most spoken
languages and is employed by an important community all over the world. Therefore, research on Vietnamese language processing is an essential task. This paper focuses on three main tasks for Vietnamese language processing including POS tagging, chunking, and NER.

In this paper, we present a state-of-the-art system namely NNVLP for the
Vietnamese language processing. NNVLP toolkit outperforms most previously published toolkits on three tasks including POS tagging, chunking, and NER. The contributions of this work consist of:
\begin{itemize}
\item We demonstrate a neural network-based system reaching
  the state-of-the-art performance for Vietnamese language processing including POS tagging, chunking, and NER. Our system is a combination of Bi-LSTM, CNN, and CRF models, which achieves an accuracy of $91.92\%$, $F_{1}$ scores of $84.11\%$ and $92.91\%$ for POS tagging, chunking, and NER tasks respectively.
\item We provide our API and web demo for user, which
  is believed to positively contributing to the long-term advancement of Vietnamese language processing.
\end{itemize}

The remainder of this paper is structured as
follows. Section~\ref{sec:relatedWork} summarizes related work on
Vietnamese language processing. Section~\ref{sec:models} describes NNVLP toolkit architecture, API, and web interface. Section~\ref{sec:experiments} gives experimental results and
discussions. Finally, Section~\ref{sec:conclusion} concludes the paper.
\section{Related Works}
\paragraph{}
Previously published systems for Vietnamese language processing used
traditional machine learning methods such as Conditional Random Field
(CRF), Maximum Entropy Markov Model (MEMM), and Support Vector Machine
(SVM). In particular, most of the toolkits for POS tagging task
attempted to use conventional models such as CRF~\cite{Mai-Vu:2013}
and MEMM~\cite{Le:2010a}.~\cite{Mai-Vu:2013} also used CRF for
chunking task. Recently, at the VLSP 2016 workshop for NER task,
several participated system use MEMM~\cite{Phuong:2016},
\cite{Van:2016} and CRF~\cite{Huong:2016} to solve this problem. 
\label{sec:relatedWork}

\section{NNVLP API and Web Demo}
\label{sec:models}
\subsection{System Architecture}
\begin{figure}[t]
\centering
\includegraphics[scale=0.35]{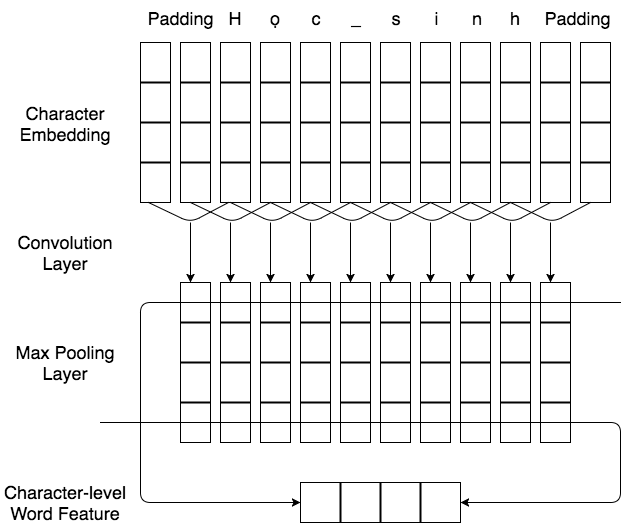}
\caption{The CNN layer for extracting character-level word features of word \textit{Học\_sinh} (Student).}
\label{fig:1}
\end{figure}

\begin{figure}[t]
\centering
\includegraphics[scale=0.15]{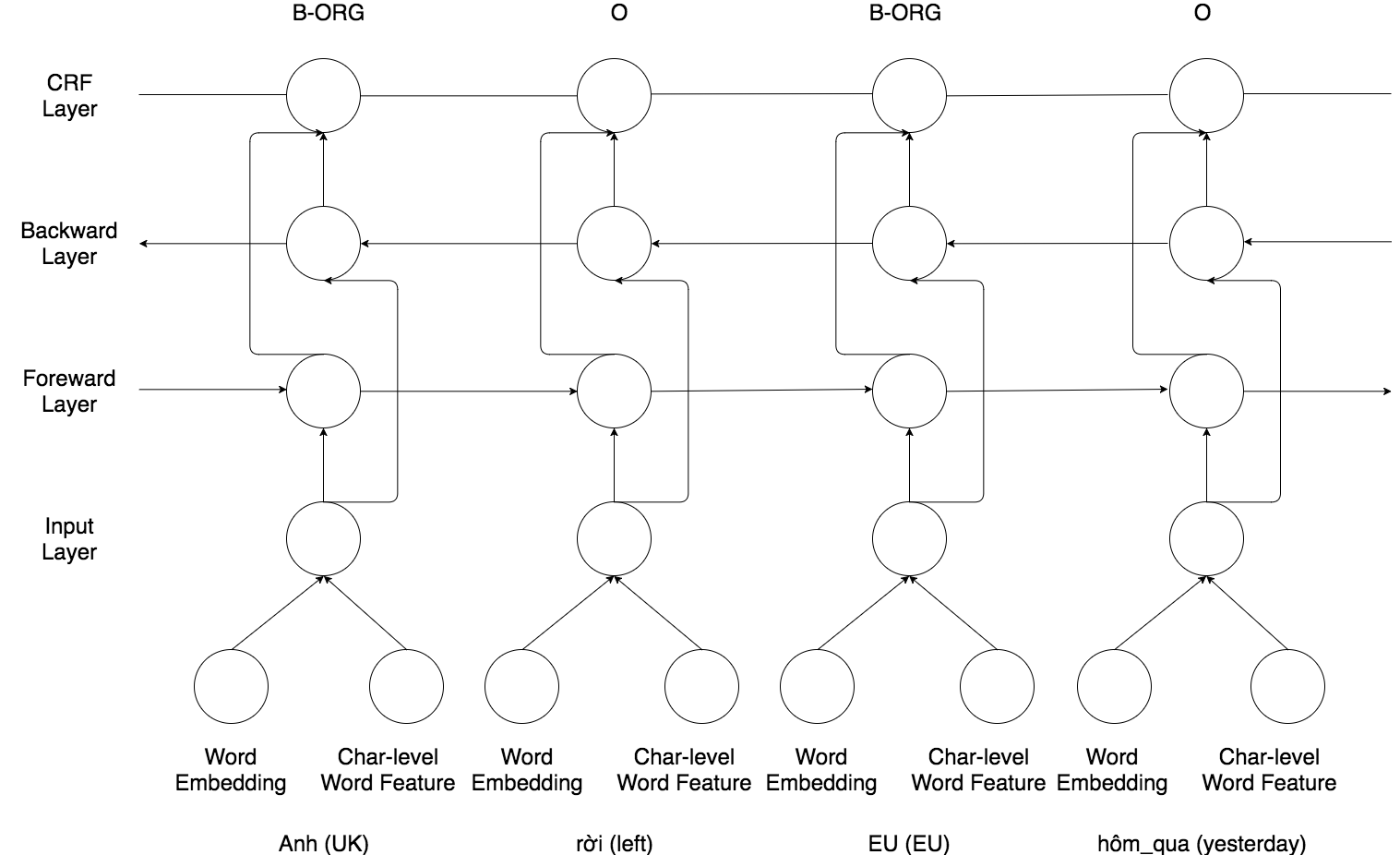}
\caption{The Bi-LSTM-CRF layers for input sentence \textit{Anh rời EU hôm
    qua.} (UK left EU yesterday.)} 
\label{fig:2}
\end{figure}

\paragraph{}
We implement the deep neural network model described
in~\cite{Thai-Hoang:2017}. This model is a combination of
Bi-directional Long Short-Term Memory (Bi-LSTM), Convolutional Neural
Network (CNN), and Conditional Random Field (CRF). In particular, this
model takes as input a sequence of the concatenation of word embedding
pre-trained by
word2vec\footnote{\url{https://code.google.com/archive/p/word2vec/}}
tool and character-level word feature trained by CNN. That sequence is 
then passed to a Bi-LSTM, and then a CRF layer takes as input the
output of the Bi-LSTM to predict the best named entity output
sequence. Figure~\ref{fig:1} and Figure~\ref{fig:2} describe the
architectures of BI-LSTM-CRF layers, and CNN layer respectively. 

\begin{figure}[h]
\centering
\includegraphics[scale=0.3]{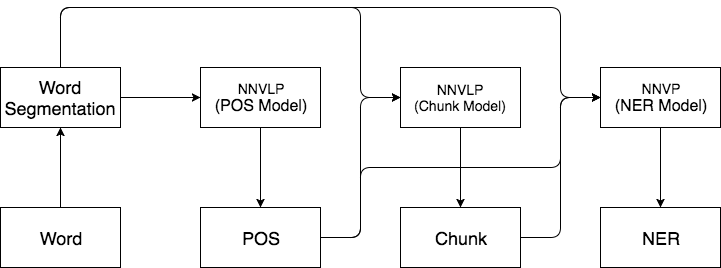}
\caption{The Architecture of NNVLP Toolkit}
\label{fig:3}
\end{figure}

NNVLP toolkit uses these architectures for all tasks including POS
tagging, chunking, and NER. Because each word in the Vietnamese
language may consist of more than one syllables with spaces in
between, which could be regarded as multiple words by the unsupervised
models, we, first, segment the input texts into sequences of words by
pyvi toolkit\footnote{\url{https://pypi.python.org/pypi/pyvi}}. These
word sequences are put into NNVLP toolkit to get corresponding POS tag
sequences. Next, these words and POS tag sequences are put into NNVLP
toolkit to get corresponding chunk sequences. Finally, NNVLP toolkit
takes as input sequences of the concatenation of word, POS tag, and
chunk to predict corresponding NER sequences. Figure~\ref{fig:3}
presents this pipeline of NNVLP toolkit. 

\subsection{NNVLP API}
\begin{figure}[h]
\centering
\includegraphics[scale=0.42]{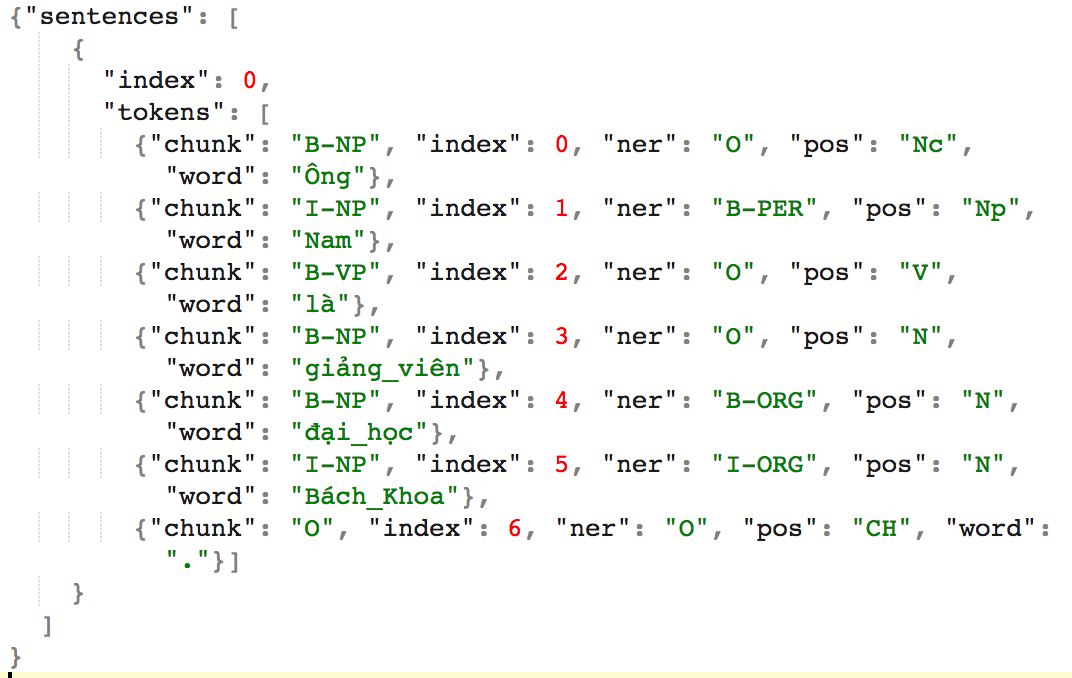}
\caption{The output JSON of the input sentence "Ông Nam là giảng viên đại học Bách Khoa." (Mr Nam is a lecturer of Bach Khoa University.)}
\label{fig:4}
\end{figure}

\paragraph{}
NNVLP API is an API for Vietnamese Language Processing which takes input 
sentences and outputs a JSON containing a list of sentences where each word in these sentences has POS tag, chunk, named entity attributes as shown in Figure~\ref{fig:4}.
\subsection{Web Demo}
\begin{figure*}[t]
\centering
\includegraphics[scale=0.255]{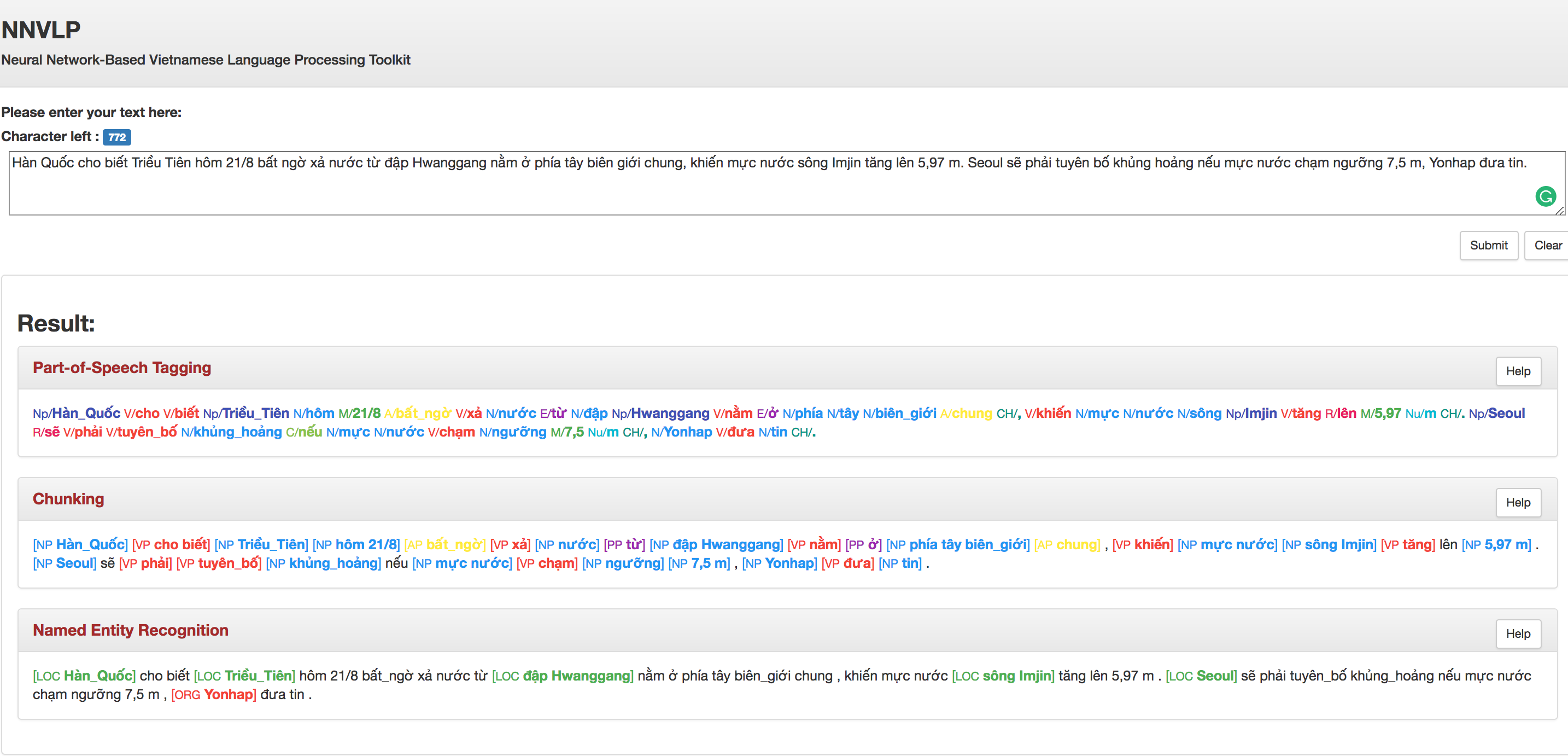}
\caption{The Web Interface of NNVLP Toolkit}
\label{fig:5}
\end{figure*}

\paragraph{}
We also provide web interface\footnote{\url{nnvlp.org}} for users of NNVLP toolkit. Users can type or paste raw texts into the textbox and click \textit{Submit} button to get the corressponding POS tag, chunk, named entity sequences. Each label is tagged with different color to make the output easy to see. Users can also look up the meaning of each label by click \textit{Help} button. Figure~\ref{fig:5} presents the web interface of our system.
\section{Experiments}
\label{sec:experiments}
\paragraph{}
In this section, we compare the performance of NNVLP toolkit with
other published toolkits for Vietnamese including
Vitk~\cite{Le:2010a}, vTools~\cite{Mai-Vu:2013},
RDRPOSTagger~\cite{nguyen-EtAl:2014:Demos}, and
vie-ner-lstm~\cite{Thai-Hoang:2017b}.

\subsection{Data Sets}
\paragraph{}
To compare fairly, we train and evaluate these systems on the VLSP
corpora. In particular, we conduct experiments on Viet Treebank corpus
for POS tagging and chunking tasks, and on VLSP shared task 2016
corpus for NER task. All of these corpora are converted to CoNLL
format. The corpus of POS tagging task consists of two columns namely
word, and POS tag. For chunking task, there are three columns namely
word, POS tag, and chunk in the corpus. The corpus of NER task
consists of four columns. The order of these columns are word, POS
tag, chunk, and named entity. While NER corpus has been separated into
training and testing parts, the POS tagging and chunking data sets are
not previously divided. For this reason, we use $80\%$ of these data
sets as a training set, and the remaining as a testing set. Because
our system adopts early stopping method, we use $10\%$ of these data
sets from the training set as a development set when training NNVLP
system. Table~\ref{tab:1} and Table~\ref{tab:2}\footnote{For more
  details about these tagsets, please visit the demo website at
  \url{nnvlp.org}} shows the statistics of each corpus.  
\begin{table}[h]
\center
{
\begin{tabular}{|c|c|c|c|}
\hline 
 & \multicolumn{3}{c|}{Number of sentences} \\ 
\hline 
Data sets & POS & Chunk & NER \\ 
\hline 
Train & 7268 & 7283 & 14861 \\ 
\hline 
Dev & 1038 & 1040 & 2000 \\ 
\hline 
Test & 2077 & 2081 & 2831 \\ 
\hline 
\end{tabular}}
\caption{The number of sentences for each part in POS tagging, chunking, and NER data sets\label{tab:1}}
\end{table}

\begin{table}[h]
\center
\resizebox{\linewidth}{!}{
\begin{tabular}{|c|c|}
\hline 
Data sets & Labels \\ 
\hline 
POS & \parbox[t]{2.1in}{N, V, CH, R, E, A, P, Np, M, C, Nc, L, T, Ny, Nu, X, B, S, I, Y, Vy} \\ 
\hline 
Chunk & NP, VP, PP, AP, QP, RP \\ 
\hline 
NER & PER, LOC, ORG, MISC \\ 
\hline 
\end{tabular}}
\caption{Labels in POS tagging, chunking, and NER data sets\label{tab:2}}
\end{table}
\subsection{Evaluation Methods}
\paragraph{}
We use the accuracy score that is the percentage of correct labels to evaluate the performance of each system for POS tagging task. For chunking and NER tasks, the performance is measured with $F_{1}$ score, where $F_1 =
\frac{2*P*R}{P + R}$. Precision ($P$) is the percentage of chunks or named entities found by
the learning system that are correct. Recall ($R$) is the percentage of chunks or
named entities present in the corpus that are found by the system. A
chunk or named entity is correct only if it is an exact match of the 
corresponding phrase in the data file.

\subsection{Experiment Results}
\paragraph{}
We evaluate performances of our system and several published systems
on POS tagging, chunking, and NER data sets. Inputs for POS tagging
task are words, for chunking task are words and POS tags, and for NER
task are words, POS tags, and chunks. Table~\ref{tab:3},
Table~\ref{tab:4}, and Table~\ref{tab:5} present the performance of
each system on POS tagging, chunking, and NER task respectively. The
hyper-parameters for training NNVLP are given in
Table~\ref{tab:params}.

\begin{table}[h]
\center
{
\begin{tabular}{|l|l|}
\hline 
System & Accuracy \\ 
\hline 
Vitk & 88.41 \\ 
\hline 
vTools & 90.73 \\ 
\hline 
RDRPOSTagger & 91.96 \\ 
\hline 
NNVLP & \textbf{91.92} \\ 
\hline 
\end{tabular}}
\caption{Performance of each system on POS tagging task\label{tab:3}}
\end{table}

\begin{table}[h]
  \centering
  \begin{tabular}{|l | l| r |}
    \hline
    Layer & Hyper-parameter & Value\\
    \hline
    CNN & window size & 3 \\
          & number of filters & 30 \\
    \hline
    LSTM & hidden nodes & 300 \\
    \hline 
    Embedding & word & 300 \\
    & character-level& 30\\
    \hline
  \end{tabular}
  \caption{Hyper-parameters of our models}
  \label{tab:params}
\end{table}

\begin{table}[h]
\center
{
\begin{tabular}{|l|l|l|l|}
\hline 
System & P & R & F1 \\ 
\hline 
vTools & 82.79 & 83.55 & 83.17 \\ 
\hline 
NNVLP & 83.93 & 84.28 & \textbf{84.11} \\ 
\hline 
\end{tabular}}
\caption{Performance of each system on chunking task\label{tab:4}}
\end{table}

\begin{table}[h]
\center
{
\begin{tabular}{|l|l|l|l|}
\hline 
System & P & R & F1 \\ 
\hline 
Vitk & 88.36 & 89.20 & 88.78 \\ 
\hline 
vie-ner-lstm & 91.09 & 93.03 & 92.05 \\ 
\hline 
NNVLP & 92.76 & 93.07 & \textbf{92.91} \\ 
\hline 
\end{tabular}}
\caption{Performance of each system on NER task\label{tab:5}}
\end{table}

By combining Bi-directional Long Short-Term Memory, Convolutional
Neural Network, and Conditional Random Field, our system outperforms
most published systems on these three tasks. In particular, NNVLP
toolkit achieves an accuracy of $91.92\%$, $F_{1}$ scores of $84.11\%$
and $92.91\%$ for POS tagging, chunking, and NER tasks respectively. 

\section{Conclusion}
\label{sec:conclusion}
We present a neural network-based toolkit for Vietnamese processing
that is a combination of Bi-LSTM, CNN, and CRF.  The system takes raw
sentences as input and produces POS tag, chunk and named entity
annotations for these sentences. The experimental results showed that NNVLP toolkit
achieves state-of-the-art results on three tasks including POS
tagging, chunking, and NER.

\bibliography{bibliography} 
\bibliographystyle{ijcnlp2017}
\end{document}